%% file: main.tex
\documentclass[10pt,twocolumn,letterpaper]{article}

\usepackage[pagenumbers]{wacv} 
\usepackage{algorithmicx}
\usepackage{algpseudocode}
\usepackage{algorithm}
\usepackage{amsmath}
\usepackage{amssymb}
\usepackage{amsfonts}

\usepackage[accsupp]{axessibility}
\usepackage{xcolor}
\usepackage{colortbl} 

\algrenewcommand\algorithmicend{}
\algrenewcommand\algorithmicdo{}
\algrenewcommand\algorithmicthen{}
\algtext*{EndFor} 
\algtext*{EndIf}  
\algtext*{EndWhile} 
\usepackage{adjustbox} 

\definecolor{wacvblue}{rgb}{0.21,0.49,0.74}
\usepackage[pagebackref,breaklinks,colorlinks,allcolors=wacvblue]{hyperref}

\def\confName{WACV}
\def\confYear{2026}

\title{HEART-PFL: Stable Personalized Federated Learning under Heterogeneity with Hierarchical Directional Alignment and Adversarial Knowledge Transfer}

\author{Minjun Kim$^{*}$ \\
Promedius Inc.\\
{\tt\small weightboy7@gmail.com}
\and
Minje Kim$^{*,\dagger}$ \\
Promedius Inc.\\
{\tt\small iankimrok@gmail.com}
}
\begin{document}
\maketitle
\renewcommand{\thefootnote}{\fnsymbol{footnote}}
\footnotetext[1]{Equal contribution}
\footnotetext[2]{Corresponding author}
\input{sec/0_abstract}    
\input{sec/1_intro}
\input{sec/2_related_works}
\input{sec/3_methodology}
\input{sec/4_experimental_results}
\input{sec/5_conclusion}
{
    \small
    \bibliographystyle{ieeenat_fullname}
    \bibliography{main}
}

\end{document}

%% file: sec/0_abstract.tex
\begin{abstract}

Personalized Federated Learning (PFL) aims to deliver effective client-specific models under heterogeneous distributions, yet existing methods suffer from shallow prototype alignment and brittle server-side distillation. We propose HEART-PFL, a dual-sided framework that (i) performs depth-aware Hierarchical Directional Alignment (HDA) using cosine similarity in the early stage and MSE matching in the deep stage to preserve client specificity, and (ii) stabilizes global updates through Adversarial Knowledge Transfer (AKT) with symmetric KL distillation on clean and adversarial proxy data. Using lightweight adapters with only 1.46M trainable parameters, HEART-PFL achieves state-of-the-art personalized accuracy on CIFAR-100, Flowers-102, and Caltech-101 (63.42\%, 84.23\%, and 95.67\%, respectively) under Dirichlet non-IID partitions, and remains robust to out-of-domain proxy data. Ablation studies further confirm that HDA and AKT provide complementary gains in alignment, robustness, and optimization stability, offering insights into how the two components mutually reinforce effective personalization. Overall, these results demonstrate that HEART-PFL simultaneously enhances personalization and global stability, highlighting its potential as a strong and scalable solution for PFL (code available at \url{https://github.com/danny0628/HEART-PFL}).
\end{abstract}

%% file: sec/1_intro.tex
\section{Introduction}
\label{sec:intro}
Federated learning (FL) is a distributed learning framework that enables multiple clients to collaboratively train models without requiring centralized data collection, thereby preserving data privacy and supporting deployment across diverse domains such as healthcare and computer vision. As a fundamental solution, the FedAvg~\cite{mcmahan2017communication} algorithm updates a global model on the server by aggregating parameters from locally trained clients. One of the central challenges in FL arises from client heterogeneity, which commonly manifests as imbalanced class distributions or non-independent and identically distributed (non-IID) data~\cite{li2019convergence, duan2023federated,xiong2023feddm,gao2022feddc}. In response to personalization under client heterogeneity, Personalized Federated Learning (PFL)~\cite{t2020personalized} couples client-specific adaptation with server-mediated knowledge aggregation to improve per-client performance.

Previous studies have investigated diverse personalization strategies. Early approaches adopted full-model personalization~\cite{hanzely2020federated, li2021ditto, mansour2020three,hanzely2020lower}, in which each client optimizes a personalized model with a regularization term while the global model resides on the server. This strategy is parameter-intensive and imposes substantial computation and communication overhead, limiting its practicality for resource-constrained clients. To alleviate the overhead, subsequent works proposed partial-model personalization~\cite{collins2021exploiting, oh2021fedbabu, arivazhagan2019fedper, zhang2023fedcp}, where each client splits its model into personalized and shared components, exchanging only shared parameters with the server. 

Adapter-based personalization~\cite{pillutla2022federated} reduces computation by personalizing only lightweight adapter modules while sharing a common backbone across clients. PerAda~\cite{xie2024perada} extends this idea by freezing all non-adapter parameters and updating only the adapter modules, enabling highly efficient personalization. 

Prototype-based approaches~\cite{tan2022fedproto,guo2025fedorgp,liu2024fedgpd} typically perform alignment using prototypes from a single layer and a predominantly server→client flow, leaving hierarchical semantics underutilized and limiting client specificity. FedPHP~\cite{li2021fedphp}, for example, depends on global prototypes whose quality mirrors that of personalized extractors—a fragile circular dependency. Meanwhile, GPFL~\cite{zhang2023gpfl} introduces trainable category embeddings that guide features in both magnitude and direction without relying on strong extractors. 

Knowledge transfer among clients via the server can guide the training process and improve personalization~\cite{zhang2021parameterized}. Incorporating clients’ historical personalized knowledge~\cite{jin2022personalized} further enhances local adaptation. However, existing knowledge distillation approaches often improve personalization at the expense of degrading the global model. To mitigate this trade-off, PerAda distills clients’ ensemble knowledge into the global model, improving generalization without sacrificing personalization.

\noindent\textbf{Motivation and Observation.}
Despite recent progress, PFL still faces two persistent gaps under non-IID client distributions and out-of-domain data. First, representation alignment is often shallow and one-way: prototype-based strategies typically align at a only single layer and mostly in the server-to-client direction, underutilizing hierarchical semantics and risking the suppression of client-specific cues essential for personalization. Second, server-side knowledge transfer remains brittle: ensemble distillation on in-distribution unseen data tends to push the global model toward an averaged client teacher in a one-sided manner, which can propagate teacher bias, fail to convey representational capacity, and destabilize global updates under heterogeneity.

\noindent\textbf{Our Approach.} We present HEART-PFL, which directly targets these two challenges with two complementary components that constitute our main contribution. Hierarchical Directional Alignment (HDA) performs layer-wise, direction-aware alignment between global features and client class prototypes: cosine-based directional agreement in early stage encourages consistent geometry for generic patterns, while MSE-based semantic matching in deep stage preserves class-specific meaning—leveraging the full hierarchy without erasing personalization (Fig.~\ref{fig:HDA}). Adversarial Knowledge Transfer (AKT) strengthens server-side transfer by making distillation bidirectional and robust. The global adapter and the client ensemble are aligned in both directions on clean and adversarial proxy views, increasing feature diversity and stabilizing updates despite client heterogeneity (Fig.~\ref{fig:akt}).
For computational and communication efficiency, we adopt an adapter-based realization following prior work~\cite{pillutla2022federated,xie2024perada}; this is a supporting design choice rather than our core contribution.

\noindent\textbf{Empirical evidence.} We evaluate HEART-PFL under standard PFL regime with label-skew, non-IID clients generated via Dirichlet sampling (various $\alpha$) and with varying client participation rates.
Across these heterogeneous settings, HEART-PFL consistently yields higher client personalization on the all benchmark datasets. Specifically, the ablation studies highlight the complementary roles of the two modules: HDA yields hierarchy-aware alignment improvements, AKT enhances robustness and server-side stability, and their combination achieves the largest personalization gains with the most stable training. We summarize our main contributions below:
\begin{itemize}
\item We propose the HEART-PFL framework, which tackles the data heterogeneity inherent in PFL through dual-side contributions: client-side alignment and server-side knowledge transfer, resulting in enhanced personalization effectiveness.
\item Our proposed method has two key components: (i) HDA, which aligns global features with personalized prototypes across multiple network layers; and (ii) AKT, which enhances the robustness of knowledge transfer via adversarial views and symmetric distillation.
\item We conduct extensive evaluations on several benchmark datasets with heterogeneous settings. Our method demonstrates superior performance compared to state-of-the-art methods across diverse heterogeneous client distributions. Moreover, we further demonstrated the effectiveness of our key components, HDA and AKT, through comprehensive ablation studies.
\end{itemize}

%% file: sec/2_related_works.tex
\section{Related Works}
\label{sec:related_works}
\subsection{Federated Learning under Data Heterogeneity}
FL often suffers significant degradation under non-IID or class-imbalanced client data, which induces client drift and destabilizes global aggregation.
The canonical FedAvg~\cite{mcmahan2017communication} struggles in such regimes, and subsequent approaches have introduced proximal constraints or variance-reduction corrections to mitigate client drift and improve convergence on skewed data~\cite{li2020fedprox,karimireddy2020scaffold}.
A complementary line of work focuses on aligning client representations via class prototypes. Methods such as FedProto~\cite{tan2022fedproto} and its successors~\cite{liu2024fedgpd,guo2025fedorgp} promote feature–prototype agreement to enhance consistency and stabilize knowledge sharing amid heterogeneous client distributions. Nonetheless, existing prototype formulations typically perform alignment at only a single feature layer and rely on a predominantly server-to-client alignment flow, which in turn underutilizes hierarchical semantic structure and attenuates client-specific characteristics.

Knowledge distillation approaches address robustness from a different angle. Server-side distillation from an ensemble of client teachers has been shown to improve the global model’s resilience to heterogeneous inputs~\cite{lin2020ensemble,lin2020feddf}. However, unilateral distillation on clean proxy data often propagates teacher bias and can destabilize global updates when clients diverge. More recent variants incorporate refined distillation strategies, yet existing approaches~\cite{yao2024fedgkd,li2019fedmd} in FL still restrict knowledge transfer to a one-way server-to-client direction. Complementary representation learning methods, including contrastive approaches~\cite{chen2020simple,li2021model}, aim to improve robustness under heterogeneity but typically operate at a single representation level and do not exploit the hierarchical structure available in deep networks. These limitations highlight the need for FL methods that can leverage multi-layer semantic information while ensuring robustness through bidirectional knowledge transfer.

\subsection{Personalized Federated Learning}

PFL aims to maximize per-client utility without forfeiting cross-client knowledge sharing. Early PFL approaches personalize the full model while regularizing toward a global reference (e.g., pFedMe~\cite{t2020pfedme}, Ditto~\cite{li2021ditto}), which improves specialization but incurs substantial computational and communication overhead. Parameter-splitting methods alleviate this burden by separating shared and private parameter subsets (e.g., FedPer~\cite{arivazhagan2019fedper}, LG-FedAvg~\cite{liang2020think}), and recent systems refine this idea through selective or adaptive update schedules (e.g., FedSelect~\cite{tamirisa2024fedselect}, FedALA~\cite{zhang2023fedala}). Adapter-based designs further reduce overhead by training compact, client-specific modules atop a shared backbone (e.g., PerAda~\cite{xie2024perada}).

Prototype-driven PFL aligns global features with client-specific prototypes, but its performance often depends heavily on extractor quality. Moreover, most formulations restrict  alignment to a single representation layer with a predominantly server-to-client flow (e.g., FedPHP~\cite{li2021fedphp}, FedProto~\cite{tan2022fedproto}). More recent efforts (e.g., FedGPD~\cite{liu2024fedgpd}, FedORGP~\cite{guo2025fedorgp}) advance this line but still center on shallow, unidirectional prototype alignment. Distillation-centric PFL, in contrast, transfers personalized knowledge through the server or historical teachers. Although modern formulations explore balanced divergence~\cite{qi2025balance} and revised KL strategies~\cite{wu2024rethinking}, most PFL applications still rely on conventional teacher–student pipelines without incorporating adversarial perturbations or symmetric constraints.

Within this landscape, our method integrates two complementary components: (i) HDA between global features and personalized class prototypes to capture multi-layer semantics while preserving client-specific information, and (ii) bidirectional, symmetric knowledge distillation on both clean and adversarial proxy views to stabilize global updates under heterogeneity. Together, these components yield improvements in both personalized and global performance across strong PFL baselines while reducing round-to-round instability.

%% file: sec/3_methodology.tex
\section{Methodology}

\subsection{Preliminaries}
\noindent\textbf{Personalized Federated Learning.}
We consider a traditional PFL setting with $N$ clients $C = \{c_1, \ldots, c_N\}$, 
where each client $c_k$ has a local non-IID dataset $\mathcal{D}_k = \{(\mathbf{x}_k^i, \mathbf{y}_k^i)\}_{i=1}^{n_k}$ 
containing $n_k$ data samples. Let $\theta \in \mathbb{R}^{d_\theta}$ denote the parameters of global model shared across all clients, 
and $\Omega = \{\omega_1, \omega_2, \ldots, \omega_{N}\}$ represent the collection of parameters of personalized models, where each $\omega_k \in \mathbb{R}^{d_\omega}$ is specific to client $c_k$. The typical objective of PFL is formulated as:
\begin{equation}
    \min_{\theta, \Omega} \mathcal{P}(\theta, \Omega) = \frac{1}{N} \sum_{k=1}^{N} \mathcal{F}_k(\theta, \omega_k),
\end{equation}
where $\mathcal{F}_k(\theta, \omega_k)$ denotes the local objective function for client $c_k$:
\begin{equation}
    \mathcal{F}_k(\theta, \omega_k) = \frac{1}{n_k} \sum_{i=1}^{n_k} \mathcal{L}(f(\mathbf{x}_k^i; \theta, \omega_k), \mathbf{y}_k^i),
\end{equation}
where $f(\cdot; \theta, \omega_k)$ is the model parameterized by both global and personalized 
parameters, and $\mathcal{L}(\cdot, \cdot)$ is the task-specific loss function.
\\\noindent\textbf{Adapter for cost efficiency.} Mitigating communication and computational overhead remains a critical challenge in PFL. To address this, we adopt a lightweight adapter architecture~\cite{xie2024perada} implemented as a sequence of BatchNorm–Dropout–Conv2D layers with a residual connection, leveraging frozen pretrained parameters to maximize efficiency. Unlike conventional formulations where both global and personalized models are fully trainable, our framework restricts trainable components to adapter modules. 

Formally, we redefine the global model as $\theta = (\psi, \tilde{\theta})$ and the personalized model as $\omega_k = (\psi, \tilde{\omega}_k)$, 
where $\psi$ denotes frozen pretrained parameters and $\tilde{\theta}, \tilde{\omega}_k$ are the trainable adapter parameters. 
The personalized objective is thus reformulated as:
\begin{equation}
    \min_{\tilde{\theta}, \tilde{\Omega}} \mathcal{P}(\tilde{\theta}, \tilde{\Omega}) = \frac{1}{N} \sum_{k=1}^{N} \mathcal{F}_k(\tilde{\theta}, \tilde{\omega}_k).
\end{equation}
\begin{figure}[htbp]
    \centering
    \includegraphics[width=0.35\textwidth]{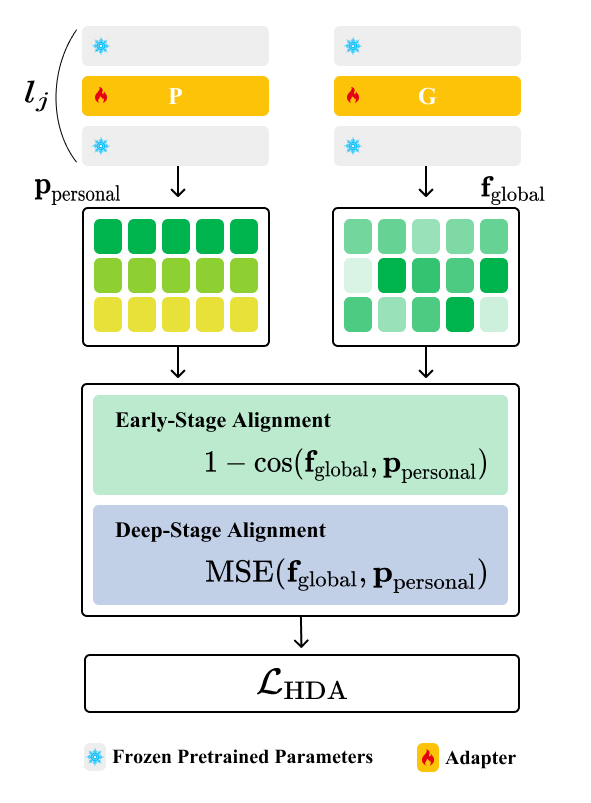}
    \caption{Overview of our proposed HDA. HDA extracts a hierarchical set of class-wise prototypes from each client's personalized model. It then aligns features from the global model with these client-specific prototypes using a semantic-aware mechanism: cosine similarity for early stage and MSE for deep stage. This alignment mechanism is formulated as our proposed HDA loss, $\mathcal{L}_{\text{HDA}}$.
}
    \label{fig:HDA}
\end{figure}
\subsection{Hierarchical Directional Alignment}
Existing prototype-based approaches~\cite{tan2022fedproto,guo2025fedorgp, liu2024fedgpd} have focused primarily on single-layer representations, which may not fully leverage the hierarchical semantic structure inherent in deep networks. Although these methods have shown promising results, they typically extract prototypes from only a single network layer, potentially overlooking information from the rest of the layers. Moreover, these methods predominantly follow a server-to-client alignment direction, where client models are regularized to conform to global prototypes. This is often at the expense of preserving the unique features required for effective personalization.

To address these challenges, we introduce HDA (Fig.~\ref{fig:HDA}), a novel layer-wise alignment mechanism that utilizes features of the global model $\theta$ with class-wise prototypes of the personalized model $\omega_k$. As illustrated in Figure~\ref{fig:HDA}, HDA captures structural relationships between global and client parameter spaces via layer-specific alignment strategies, thereby enhancing personalized performance for all clients. \\
\noindent\textbf{Layer-wise Prototype Extraction.}  For each client $c_k$, we extract multi-scale feature representations across 
$l$ network layers, spanning from elementary feature to semantic features. 
Given the client dataset $\mathcal{D}_k$, we formalize the layer-wise feature extraction process as:

\begin{equation}\mathbf{F}^{(l)}_k = \{h^{(l)}_k(\mathbf{x}_i) : (\mathbf{x}_i, \mathbf{y}_i) \in \mathcal{D}_k\},
\end{equation}
where $h$ denotes the feature extraction function, followed by adaptive pooling to ensure dimensional consistency. 
\\\noindent\textbf{Class-wise Prototype Extraction.} For each class $v$  and layer $l$, we construct personalized prototypes by aggregating class-representative features:

\begin{equation}
\label{eq:prototype_update}
\mathbf{p}^{(l)}_{k,v} = \frac{1}{|\mathcal{D}_{k,v}|} \sum_{(\mathbf{x}_i, \mathbf{y}_i) \in \mathcal{D}_{k,v}} h^{(l)}_k(\mathbf{x}_i).
\end{equation}
This generates a hierarchical prototype collection
$\mathbf{P}_k$, 
capturing abundant class representations for client $k$.
The class-wise prototype extraction is consistent across heterogeneous data distributions, aiding reliable convergence within each client’s imbalanced class distribution.

\noindent\textbf{Semantic-Aware Layer-wise Alignment.} The core innovation of HDA lies in its semantic-aware alignment mechanism that 
employs stage-wise similarity functions tailored to different representational characteristics. 
The semantic-aware alignment function is defined as:
\begin{equation}
\mathcal{A}^{(l)}=
\begin{cases}
1 - \cos(\mathbf{f}_{\text{global}},\, \mathbf{p}_{\text{personal}}), & l \in l_{\text{early}},\\[4pt]
\mathrm{MSE}(\mathbf{f}_{\text{global}},\, \mathbf{p}_{\text{personal}}), & l \in l_{\text{deep}}.
\end{cases}
\end{equation}

Our asymmetric design leverages global features $\mathbf{f}_{\text{global}}$ extracted by the global model and aligns them with client-specific, class-wise prototypes $\mathbf{p}_{\text{personal}}$, effectively bridging client-specific understanding with collective global knowledge.
Here, $l_{\text{early}}$ and $l_{\text{deep}}$ denote the layer sets of the early stage and the deep stage, respectively. We partition layers by depth: the early stage spans from the input to the network’s central region, and the deep stage covers the layers beyond this boundary, as validated by our ablation study (Figure~\ref{fig:HDA_ablation}).  In the early stage, where features are more generic, we employ cosine similarity to guide their directional alignment. For the deep stage, we use Euclidean distance to enforce precise alignment.

The complete training objective for each client combines the standard cross-entropy loss, proximal term $\frac{1}{2}\|\omega_k - \theta\|_2^2$, and our proposed HDA loss:
\begin{equation}
\mathcal{L}_{\text{HDA},k}(\theta, \mathbf{P}_k) = \frac{1}{n_k} \sum_{i=1}^{n_k} \sum_{j=1}^{l} \frac{1}{l} \mathcal{A}^{(j)}(h^{(j)}_g(\mathbf{x}_k^i; \theta), \mathbf{p}^{(j)}_{k,\mathbf{y}_k^i}).
\end{equation}
\noindent Consequently, the final personalized objective function is:
\begin{equation}
\label{eq:HDA_final_loss}
    \min_{\theta, \Omega} \mathcal{P}(\theta, \Omega) = \frac{1}{N} \sum_{k=1}^{N} \left\{ \mathcal{F}_k(\theta, \omega_k) + \mathcal{L}_{\text{HDA},k}(\theta, \mathbf{P}_k) \right\}.
\end{equation}
\begin{figure}[htbp]
    \centering
    \includegraphics[width=0.35\textwidth]{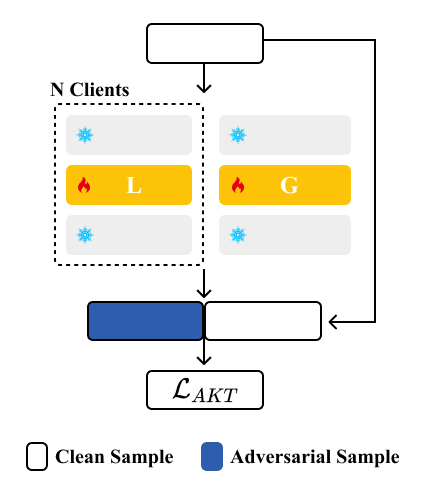}
    \caption{Overview of our proposed AKT. To enhance the robustness of the global adapter, AKT performs knowledge distillation using both clean and adversarially generated proxy samples.}
    \label{fig:akt}
\end{figure}
\subsection{Adversarial Knowledge Transfer}
We propose an AKT mechanism that improves the generalization of the global adapter by enhancing the robustness of ensemble Knowledge Transfer (EKT)~\cite{lin2020ensemble,xie2024perada} from heterogeneous clients to the server. Beyond aggregating client-specific parameters, our AKT further distills the client-ensemble knowledge to update the global adapter, thereby ensuring stable convergence under client heterogeneity. 

On the server side, EKT employs a proxy dataset $\mathcal{D}_{\rho}$ to minimize the discrepancy between the probability distributions of the clients’ local models and the global model:
\begin{equation}
\mathcal{L}_{EKT}
=\sum_{\mathbf{x}_\rho\in \mathcal{D}_{\rho}}
\mathcal{L}_{KL}\!\left(\bar{p}(\mathbf{x}_\rho)\,\|\,p_g(\mathbf{x}_\rho)\right),
\end{equation}

where $\mathcal{L}_{KL}(\cdot\,\|\,\cdot)$ is the Kullback–Leibler divergence (KL) loss~\cite{hinton2015distilling}.
Here, $\bar{p}(\mathbf{x}_\rho)=\frac{1}{N}\sum_{i=1}^{N} p_i(\mathbf{x}_\rho)$ denotes the ensemble (averaged) client distribution,
and $p_g(\mathbf{x}_\rho)$ denotes the probability distribution of the global model; both are obtained by applying the softmax to their corresponding logits.

However, this unilateral knowledge distillation from the local ensemble to the global model may lead to superficial transfer of knowledge. To address this, we enhance the EKT framework by incorporating adversarial knowledge and symmetric distillation, which promote feature diversity through adversarial samples and improve the stability of global adapter updates.

We utilize the symmetric KL loss, inspired by~\cite{wu2024rethinking,qi2025balance,lee2023self}, to reduce representational discrepancies between the global adapter and the per-client local adapters under data heterogeneity. Specifically, for each client $k$, we instantiate a per-client local adapter $\tilde{\phi}_k$, used solely for EKT to the global adapter; $\tilde{\phi}_k$ mirrors the adapter architecture of both the personalized and global models. The symmetric KL acts as a regularizer, encouraging the global adapter to better capture the representational capacity of the local adapters while avoiding one-sided overfitting. Formally, the symmetric KL loss is defined as follows:
\begin{equation}
\begin{aligned}
\mathcal{L}_{sKL}
=\,&
\sum_{\mathbf{x}_\rho\in\mathcal{D}_{\rho}}
\mathcal{L}_{KL}\big(\bar{p}(\mathbf{x}_\rho)\,\|\,p_g(\mathbf{x}_\rho)\big) \\
&+
\sum_{\mathbf{x}_\rho\in\mathcal{D}_{\rho}}
\mathcal{L}_{KL}\big(p_g(\mathbf{x}_\rho)\,\|\,\bar{p}(\mathbf{x}_\rho)\big).
\end{aligned}
\end{equation}

Then, we employ the standard adversarial perturbation method~\cite{madry2017towards} to generate adversarial samples that embed robust knowledge. Given a clean sample $(\mathbf{x}_\rho, \mathbf{y}_\rho)$ from the proxy dataset $\mathcal{D}_{\rho}$, the adversarial counterpart $\mathbf{x}_\rho^{adv}$ is obtained via:
\begin{equation}
\label{eq:pgd_update}
\mathbf{x}^{t+1}_\rho = \Pi_{\mathcal{B}_\epsilon(\mathbf{x}_\rho)}\Big(
\mathbf{x}^t_\rho + \alpha \cdot \mathrm{sign}\!\big(\nabla_{\mathbf{x}^t_\rho}
\mathcal{L}_{CE}(f(\mathbf{x}^t_\rho), \mathbf{y}_\rho)\big)\Big),
\end{equation}
where $\mathcal{L}_{CE}(\cdot)$ is the cross-entropy loss, $f$ is the model, $\alpha$ is the step size, and $\Pi_{\mathcal{B}_\epsilon(\mathbf{x}_\rho)}(\cdot)$ projects onto the $\ell_\infty$-ball of radius $\epsilon$ centered at $\mathbf{x}_\rho$. The process is initialized with $\mathbf{x}^0_\rho = \mathbf{x}_\rho + \delta$, where $\delta \sim \mathrm{Unif}[-\epsilon, \epsilon]$. After $T$ iterations, the final adversarial sample is $\mathbf{x}_\rho^{adv} = \mathbf{x}_\rho^T$.

Building upon the adversarial sample generation, we define the AKT loss as follows:
\begin{equation}
\mathcal{L}_{AKT}=\sum_{\mathbf{x}^{\prime}_\rho\in \{\mathbf{x}_\rho,\mathbf{x}_\rho^{adv}\} }\mathcal{L}_{sKL}(\mathbf{x}^{\prime}_\rho).
\end{equation}

\subsection{Algorithm}
\newcommand{\Input}{\item[\textbf{Input:}]}
\newcommand{\Output}{\item[\textbf{Output:}]}
\renewcommand{\algorithmicindent}{2em}
\begin{algorithm}
\caption{HEART-PFL}
\label{alg:HEART-PFL_algo}
\begin{algorithmic}[1]
\Input Client set $C$, rounds $R$, Initialized models ${\theta}^0$,$\{\tilde{\omega}_k^{0}\}_{k\in C}$, client datasets $\{\mathcal{D}_k\}_{k\in C}$, proxy dataset $\mathcal{D}_{\rho}$
\Output personalized adapters $\{\tilde{\omega}_{k}^{R}\}_{k\in C}$
\For{each round $r = \{0, 1, \ldots, R-1\}$} \textbf{do}
    \State $S_r \leftarrow$ Select a random subset of clients from $C$
    \State Receive global adapter $\tilde{\theta}^r$ from server 
    \State {\color{teal}\textbf{// Client Side}}   
    \For{client $c_k \in S_r$} \textbf{do}
        \For{epoch $e \in \{0, 1, \ldots, E_{\text{client}}-1\}$} \textbf{do}
   
            \State Update $\tilde{\omega}_k^{r,e+1}$ by Eq.~(\ref{eq:HDA_final_loss})
        \EndFor
        \State $\tilde{\omega}_k^{r+1} \leftarrow \tilde{\omega}_k^{r,E_{\text{client}}}$
        \State Initialize local adapter $\tilde{\phi}_k^{r,0} \leftarrow \tilde{\theta}^r$
        \For{epoch $e \in \{0, 1, \ldots, E_{\text{client}}-1\}$} \textbf{do}
\State $\tilde{\phi}_k^{r,e+1} \leftarrow \tilde{\phi}_k^{r,e} - \eta_{\tilde\phi} \nabla \mathcal{L}_{CE}(\psi, \tilde{\phi}_k^{r,e})$
        \EndFor
        \State \textbf{Return} ${\tilde\phi}_k^{r+1} = {\tilde\phi}_k^{r,E_{\text{client}}}$ to server
    \EndFor
    \State {\color{purple}\textbf{// Server Side}}
    \State Adapter Averaging: $\tilde{\theta}^{r+1} \leftarrow \frac{1}{|S_r|}\sum_{c_k \in S_r}\tilde\phi_k^{r+1}$
    \For{each $(\mathbf{x}_\rho, \mathbf{y}_\rho) \in \mathcal{D}_{\rho}$} \textbf{do}
    
            \State $\mathbf{x}^{adv}_{\rho,k} \gets \text{generate by Eq.~\eqref{eq:pgd_update}}   (\mathbf{x}_\rho;\, \phi_k^{r+1})$
        
            \State $\mathbf{x}^{adv}_{\rho,g} \gets \text{generate by Eq.~\eqref{eq:pgd_update}}   (\mathbf{x}_\rho;\, {\theta}^{r+1})$

    \EndFor
    \For{each $(\mathbf{x}_\rho, \mathbf{y}_\rho) \in \mathcal{D}_{\rho}$} \textbf{do}

        \State $\tilde{\theta}^{r+1} \gets \tilde{\theta}^{r+1}-\eta_{\tilde\theta} \nabla \mathcal{L}_{AKT}(\mathbf{x}^{\prime}_\rho)$
    \EndFor

    \State \textbf{Return} updated global adapter $\tilde{\theta}^{r+1}$ to all clients
\EndFor
\end{algorithmic}
\end{algorithm}
Algorithm~\ref{alg:HEART-PFL_algo} outlines the optimization of the personalized objective across rounds.
At each communication round $r \in \{0, \ldots, R-1\}$, the server samples a client subset $S_r\in C$ and broadcasts the current global adapter $\tilde{\theta}^{r}$ to those clients.
Each selected client $c_k\in S_r$ first updates its personalized adapter on local data $\mathcal{D}_k$ by minimizing the final personalized objective loss $\mathcal{P}(\cdot,\cdot)$ (Eq.~\ref{eq:HDA_final_loss}) using HDA for client epochs $\tilde{\omega}_k^{r+1}\leftarrow\tilde{\omega}_k^{r,E_{\text{client}}}$. Then,
the client initializes local adapter $\tilde{\phi}_k^{r,0} \leftarrow \tilde{\theta}^r$, and uses it with the standard cross-entropy loss $\mathcal{L}_{CE}(\psi,\tilde{\phi}_k)$ using learning rate $\eta_{\tilde{\phi}}$ over client epochs $E_{\text{client}}$ to obtain $\tilde{\phi}_k^{r+1}$, which is sent back to the server.
After collecting ${\tilde{\phi}_{k}^{r+1}}$, the server performs simple adapter averaging,
$\tilde{\theta}^{r+1} \leftarrow \frac{1}{|S_r|}\sum_{c_k \in S_r}\tilde\phi_k^{r+1}$.
To strengthen $\tilde{\theta}^{r+1}$ against client heterogeneity via AKT, the server uses the proxy dataset $\mathcal{D}_{\rho}$ to generate adversarial variants via adversarial perturbation steps with respect to the local adapters and the current global adapter (Eq.~\ref{eq:pgd_update}), producing $\mathbf{x}^{\text{adv}}_{\rho,k}$ and $\mathbf{x}^{\text{adv}}_{\rho,g}$. Next, the server updates the global adapter on the combined samples $\mathbf{x}'_\rho$ by minimizing the AKT loss $\mathcal{L}_{\text{AKT}}$ with learning rate $\eta_{\theta}$. Finally, the server broadcasts the updated adapter to initiate round $r+1$. After $R$ rounds, HEART-PFL returns the personalized adapters $\{\tilde{\omega}_{k}^{R}\}_{k\in C}$.

%% file: sec/4_experimental_results.tex
\begin{table*}[t!]
\centering
{\fontsize{9}{10}\selectfont
\renewcommand{\arraystretch}{1.2}
\setlength{\tabcolsep}{4.5pt}
\begin{tabular}{@{}lcccccccccc@{}}
\toprule
Method
& \multicolumn{3}{c}{CIFAR100} 
& \multicolumn{3}{c}{Flowers102} 
& \multicolumn{3}{c}{Caltech101} \\
\cmidrule(lr){2-4} \cmidrule(lr){5-7} \cmidrule(lr){8-10}
& $\alpha=0.1$ & $\alpha=0.3$ & $\alpha=0.5$ 
& $\alpha=0.1$ & $\alpha=0.3$ & $\alpha=0.5$ 
& $\alpha=0.1$ & $\alpha=0.3$ & $\alpha=0.5$ \\ 
\midrule
FedAvg-per   & 45.64{\scriptsize$\ \pm\ $0.4} & 45.44{\scriptsize$\ \pm\ $0.5} & 45.39{\scriptsize$\ \pm\ $0.2}
             & 68.11{\scriptsize$\ \pm\ $1.4} & 72.68{\scriptsize$\ \pm\ $0.3} & 72.77{\scriptsize$\ \pm\ $0.6}
             & 90.03{\scriptsize$\ \pm\ $0.8} & 91.89{\scriptsize$\ \pm\ $0.3} & 92.68{\scriptsize$\ \pm\ $0.7} \\

FedProx-per  & 45.83{\scriptsize$\ \pm\ $0.9} & 45.17{\scriptsize$\ \pm\ $0.6} & 45.50{\scriptsize$\ \pm\ $0.5}
             & 68.05{\scriptsize$\ \pm\ $2.1} & 72.97{\scriptsize$\ \pm\ $0.2} & 73.50{\scriptsize$\ \pm\ $0.9}
             & 89.79{\scriptsize$\ \pm\ $1.0} & 91.73{\scriptsize$\ \pm\ $0.3} & 92.63{\scriptsize$\ \pm\ $0.8} \\

FedProto-per & 38.89{\scriptsize$\ \pm\ $13.3} & 41.05{\scriptsize$\ \pm\ $0.5} & 36.03{\scriptsize$\ \pm\ $0.6}
             & 76.87{\scriptsize$\ \pm\ $0.7}  & 63.06{\scriptsize$\ \pm\ $1.2} & 57.29{\scriptsize$\ \pm\ $0.8}
             & 90.77{\scriptsize$\ \pm\ $1.1}  & 86.39{\scriptsize$\ \pm\ $0.9} & 84.09{\scriptsize$\ \pm\ $0.5} \\

FedPer       & 56.65{\scriptsize$\ \pm\ $0.3} & 61.19{\scriptsize$\ \pm\ $0.4} & 56.74{\scriptsize$\ \pm\ $0.3}
             & 81.31{\scriptsize$\ \pm\ $0.8} & 73.96{\scriptsize$\ \pm\ $0.6} & 72.08{\scriptsize$\ \pm\ $1.2}
             & 93.11{\scriptsize$\ \pm\ $0.7} & 89.07{\scriptsize$\ \pm\ $0.4} & 88.03{\scriptsize$\ \pm\ $0.3} \\

LG-FedAvg    & 43.33{\scriptsize$\ \pm\ $0.3} & 50.04{\scriptsize$\ \pm\ $0.4} & 43.69{\scriptsize$\ \pm\ $0.2}
             & 76.51{\scriptsize$\ \pm\ $1.1} & 61.63{\scriptsize$\ \pm\ $1.5} & 55.41{\scriptsize$\ \pm\ $0.4}
             & 91.53{\scriptsize$\ \pm\ $0.1} & 86.07{\scriptsize$\ \pm\ $0.3} & 83.57{\scriptsize$\ \pm\ $0.8} \\

Ditto        & 57.31{\scriptsize$\ \pm\ $9.1} & 48.72{\scriptsize$\ \pm\ $0.4} & 45.35{\scriptsize$\ \pm\ $0.2}
             & 71.81{\scriptsize$\ \pm\ $1.3} & 72.92{\scriptsize$\ \pm\ $0.3} & 73.24{\scriptsize$\ \pm\ $0.7}
             & 90.08{\scriptsize$\ \pm\ $1.0} & 91.92{\scriptsize$\ \pm\ $0.3} & 92.51{\scriptsize$\ \pm\ $0.8} \\

FedBABU      & 49.66{\scriptsize$\ \pm\ $2.4} & 48.40{\scriptsize$\ \pm\ $0.5} & 45.53{\scriptsize$\ \pm\ $0.2}
             & 68.37{\scriptsize$\ \pm\ $1.7} & 73.30{\scriptsize$\ \pm\ $0.2} & 73.66{\scriptsize$\ \pm\ $0.6}
             & 91.50{\scriptsize$\ \pm\ $0.7} & 91.90{\scriptsize$\ \pm\ $1.2} & 92.01{\scriptsize$\ \pm\ $0.7} \\

FedALA       & 51.44{\scriptsize$\ \pm\ $0.6} & 54.46{\scriptsize$\ \pm\ $0.5} & 51.20{\scriptsize$\ \pm\ $0.4}
             & 68.63{\scriptsize$\ \pm\ $2.1} & 72.05{\scriptsize$\ \pm\ $0.4} & 72.64{\scriptsize$\ \pm\ $1.4}
             & 90.86{\scriptsize$\ \pm\ $0.8} & 92.02{\scriptsize$\ \pm\ $0.7} & 92.42{\scriptsize$\ \pm\ $0.8} \\

PerAda       & 62.44{\scriptsize$\ \pm\ $0.4} & 61.42{\scriptsize$\ \pm\ $0.6} & 60.40{\scriptsize$\ \pm\ $0.6}
             & 80.83{\scriptsize$\ \pm\ $0.6} & 81.55{\scriptsize$\ \pm\ $0.5} & 81.55{\scriptsize$\ \pm\ $0.6}
             & 90.24{\scriptsize$\ \pm\ $0.8} & 87.14{\scriptsize$\ \pm\ $4.5} & 89.68{\scriptsize$\ \pm\ $0.9} \\

\midrule
\textbf{HEART-PFL} & 
\textbf{63.42{\scriptsize$\ \pm\ $0.1}} & 
\textbf{61.47{\scriptsize$\ \pm\ $0.2}} & 
\textbf{61.28{\scriptsize$\ \pm\ $0.4}} &
\textbf{84.07{\scriptsize$\ \pm\ $0.5}} & 
\textbf{83.68{\scriptsize$\ \pm\ $0.1}} & 
\textbf{84.23{\scriptsize$\ \pm\ $0.2}} &
\textbf{95.67{\scriptsize$\ \pm\ $0.3}} & 
\textbf{94.49{\scriptsize$\ \pm\ $0.3}} & 
\textbf{94.35{\scriptsize$\ \pm\ $0.2}} \\

\bottomrule
\end{tabular}
}
\caption{Personalized test accuracy (\%) of HEART-PFL and state-of-the-art baselines on CIFAR100, Flowers102, and Caltech101 under Dirichlet client partitions ($\alpha \in \{0.1, 0.3, 0.5\}$). Results are reported as mean$\pm$std over three seeds.}
\label{tab:sota_comparison}
\end{table*}

\section{Experimental Results}

\subsection{Experimental Setup}
\noindent\textbf{Datasets and Model.} 
We conduct a classification task using three datasets: CIFAR100~\cite{krizhevsky2009learning}, Flowers102~\cite{nilsback2008automated}, and Caltech101~\cite{fei2007learning}. We perform experiments in a PFL environment with simulation settings. To simulate imbalanced data distributions across all three datasets, we create non-IID environments by employing Dirichlet distributions with various alpha parameters (0.1, 0.3, 0.5) across all clients.
For the overall experiments, we adopt ResNet-18 pretrained on ImageNet-1K~\cite{russakovsky2015imagenet}, following previous studies~\cite{xie2024perada, gao2022feddc,tamirisa2024fedselect,pillutla2022federated}.
 
\noindent\textbf{Baselines.} To evaluate our method, we compare against state-of-the-art PFL methods: LG-FedAvg~\cite{liang2020think}, FedPer~\cite{arivazhagan2019fedper}, Ditto~\cite{li2021ditto}, FedBABU~\cite{oh2021fedbabu}, FedALA~\cite{zhang2023fedala}, and PerAda~\cite{xie2024perada}. For a fair comparison in personalized experiments, we include FedAvg~\cite{mcmahan2017communication}, FedProx~\cite{li2020federated}, and FedProto~\cite{tan2022fedproto}, which are originally designed for generalized global model learning, diverging from PFL's personalization objectives. Furthermore, we evaluate personalized versions of these methods, derived by fine-tuning their global models. We denote them as FedAvg-Per, FedProx-Per, and FedProto-Per, respectively.

\noindent\textbf{Implementation Details.} For fair non-IID settings, we apply the following configuration across all participating clients per round and datasets. The total number of clients $N$ is set to $20$, with 8 clients randomly participating in each round. Each round consists of client epochs $E=10$, and we conduct experiments for a total round $R=200$. For client-side hyper-parameters, we use SGD optimizer with a learning rate of $0.01$, learning rate decay of $1$, and training batch size of 16. For server-side hyper-parameters in AKT, we employ Adam optimizer with a learning rate of $0.001$ and batch size of $2048$.

\subsection{Personalization Performance Evaluation.}
Table~\ref{tab:sota_comparison} summarizes the final experimental results, comparing our proposed method, HEART-PFL, with several state-of-the-art baselines across three public benchmark datasets. 
The experiments are conducted under various levels of data heterogeneity, simulated using a Dirichlet distribution. 
The results demonstrate that HEART-PFL consistently achieves the best performance across all datasets and under every tested Dirichlet setting. 
Specifically, when averaging the performance across all distributions, HEART-PFL demonstrates substantial improvements over various baseline categories. Compared to generic FL methods adapted for personalization, HEART-PFL achieves improvements of 16.56\% on CIFAR100, 12.65\% on Flowers102, and 3.38\% on Caltech101. Against PFL methods with different approaches from HEART-PFL, our method shows gains of 11.14\%, 12.83\%, and 4.40\% across the three datasets. When compared to feature alignment-based PFL methods, HEART-PFL outperforms by margins of 23.40\%, 18.25\%, and 7.75\% respectively. Furthermore, against EKT-based PFL, our approach achieves superior performance with improvements of 0.64\%, 2.68\%, and 5.82\% across the datasets.
From an efficiency perspective, Table~\ref{tab:params} demonstrates that HEART-PFL requires lower training costs while achieving the highest performance.

\begin{table}[t!]
\centering
{\fontsize{9}{10}\selectfont
\setlength{\tabcolsep}{6pt}
\begin{tabular}{@{}lcc@{}}
\toprule
Method & Trainable Params \\
\midrule
FedAvg-per   & 11.18M  \\
FedProx-per  & 11.18M  \\
FedProto-per & 11.18M  \\ 
FedPer       & 11.18M  \\
LG-FedAvg    & 11.18M  \\
Ditto        & 22.36M  \\
FedBABU      & 11.18M  \\
FedALA       & 11.18M  \\
PerAda       & 2.82M   \\
\midrule
\textbf{HEART-PFL} & \textbf{1.46M} \\
\bottomrule
\end{tabular}
}

\caption{Comparison of trainable parameter counts across state-of-the-art PFL baselines under an identical backbone setting. HEART-PFL achieves the lowest parameter cost (1.46M)}
\label{tab:params}
\end{table}
\subsection{Ablation Study}
\noindent\textbf{Component-wise Evaluation in HEART-PFL.} To validate the effectiveness of our proposed method, we conduct comprehensive ablation studies on its two key components: HDA and AKT (Table~\ref{tab:ablation1_table}). We systematically evaluate the impact of each component on both model performance and the number of trainable parameters. Our approach achieves the highest personalized test accuracy of 63.42\% with only 1.46M adapter parameters, demonstrating that HDA and AKT each contribute substantially to the overall performance and that their combination yields optimal results in terms of both accuracy and parameter efficiency.

\begin{table}[t!]
\centering
\resizebox{\columnwidth}{!}{%
\begin{tabular}{lcc}
\toprule
Method & Trainable Params & Accuracy (\%) \\
\midrule
Baseline      & 11.18M & 45.64{\scriptsize$\ \pm\ $0.4} \\ 
+ HDA         & 11.18M & 58.94{\scriptsize$\ \pm\ $0.5} \\
+ AKT         & 11.18M & 59.46{\scriptsize$\ \pm\ $0.2} \\
+ HDA + AKT   & 11.18M & 61.83{\scriptsize$\ \pm\ $0.5} \\
\midrule
\textbf{HEART-PFL (w/ adapter)} 
              & \textbf{1.46M} 
              & \textbf{63.42{\scriptsize$\ \pm\ $0.5}} \\
\bottomrule
\end{tabular}
}
\caption{Component-wise ablation study for HEART-PFL.}
\label{tab:ablation1_table}
\end{table}

\noindent\textbf{Out-of-Domain Robustness of HEART-PFL.} To evaluate the domain generalization ability of HEART-PFL, we assess its performance when the proxy dataset $\mathcal{D}_{\rho}$ and the client datasets $\mathcal{D}_k$ originate from different domains. We consider two experimental conditions: (i) distillation using out-of-domain proxy data and (ii) distillation using in-domain proxy data. Under each condition, we further examine two scenarios.

\begin{figure}[t!]
    \centering
    \subcaptionbox{Personalized test accuracy on the CIFAR100 client dataset, using Flowers102 as the out-of-domain proxy dataset.\label{fig:fedas-cifar}}[0.49\linewidth]{
        \includegraphics[width=\linewidth,height=4.5cm,keepaspectratio]{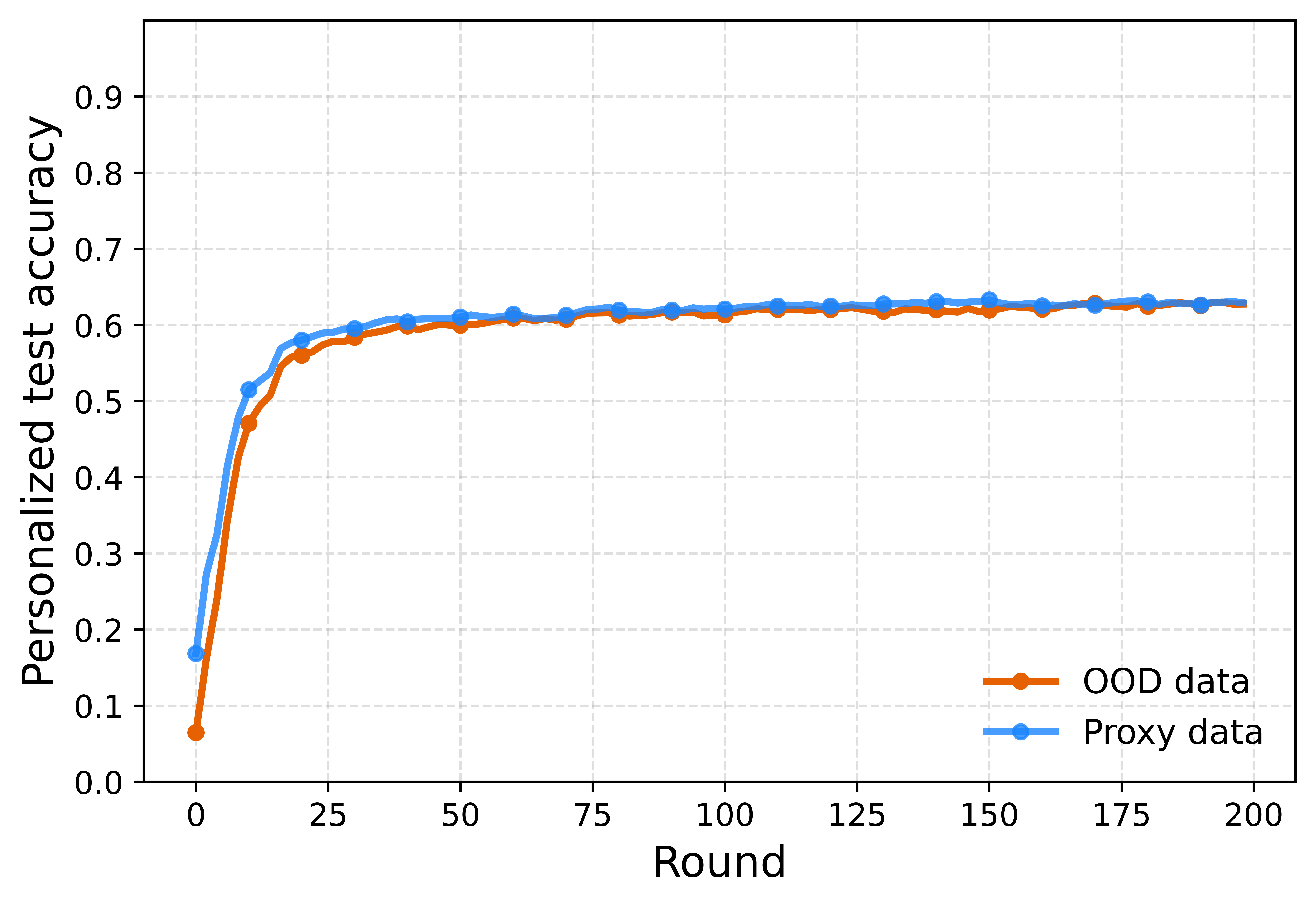}}
    \hfill
    \subcaptionbox{Personalized test accuracy on the Caltech101 client dataset, using Flowers102 as the out-of-domain proxy dataset.\label{fig:ood-cifar}}[0.49\linewidth]{
        \includegraphics[width=\linewidth,height=4.5cm,keepaspectratio]{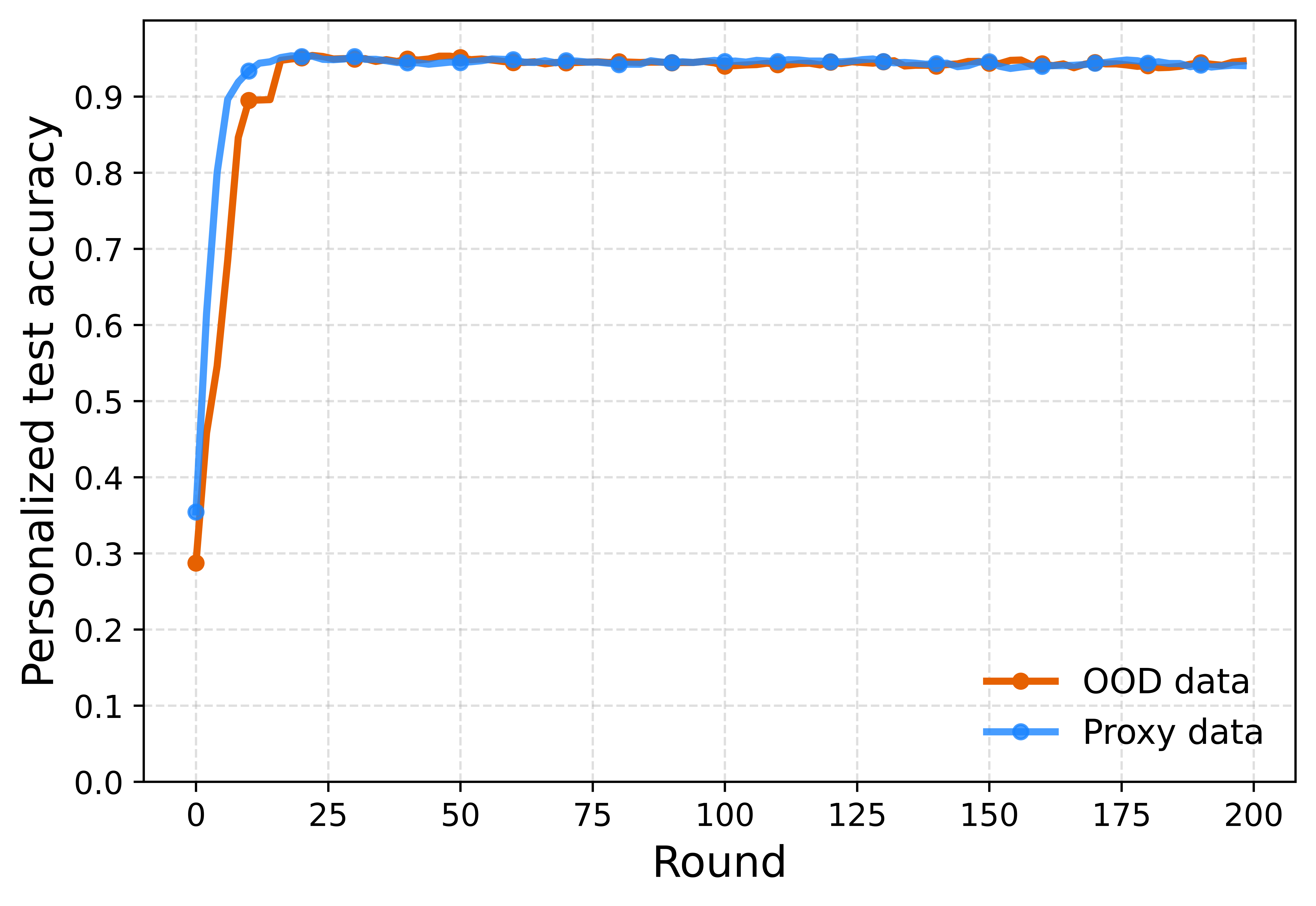}}
    \caption{Out-of-Domain Setting on CIFAR100 and Caltech101. These experiments were conducted using the methods from our HEART-PFL to measure the out-of-domain performance of AKT.}
\end{figure}

In the first scenario (Figure~\ref{fig:fedas-cifar}), the client dataset is CIFAR100 and out-of-domain proxy dataset is Flowers102, our method demonstrates remarkable domain generalization capability. The out-of-domain version achieves 63.01\% accuracy while the in-domain version reaches 63.42\% accuracy, showing only a marginal 0.41\% difference. This minimal gap indicates that HEART-PFL effectively transfers knowledge even across significantly different domains.

In the second scenario (Figure~\ref{fig:ood-cifar}), with Caltech101 as the client dataset and Flowers102 as the out-of-domain proxy dataset, we observe even more compelling results. The out-of-domain version achieves 95.40\% accuracy, while the in-domain version obtains 95.35\% accuracy. In this case, the out-of-domain approach actually outperforms the in-domain method. The HEART-PFL method demonstrates comparable performance regardless of whether the proxy dataset is in-domain or out-of-domain.

\noindent\textbf{Stage-wise HDA Design.} Our HDA method aligns personalized prototypes with global features using a dual similarity scheme: cosine similarity in early stage for directional alignment and MSE in deep stage for semantic alignment. We conduct an ablation study to verify whether this approach ensures optimal alignment at the appropriate stage. 

To analyze our method, we employ two metrics. 
First, we measure Representation Alignment, defined as the cosine similarity between global features and personalized prototypes. Cosine similarity is a primary metric for evaluating semantic alignment~\cite{chen2020simple, li2021model}, which, in our HDA, directly validates its effectiveness in achieving directional consistency. Second, we compute Feature Norm Variance, the standard deviation of feature $\ell_2$ norms, measuring how differently the model recognizes and processes various inputs to assess its representation capacity. This is motivated by findings that feature norm distributions encode meaningful model and data characteristics~\cite{park2023understanding}.

\begin{figure}[t!]
    \centering

    \includegraphics[width=\linewidth]{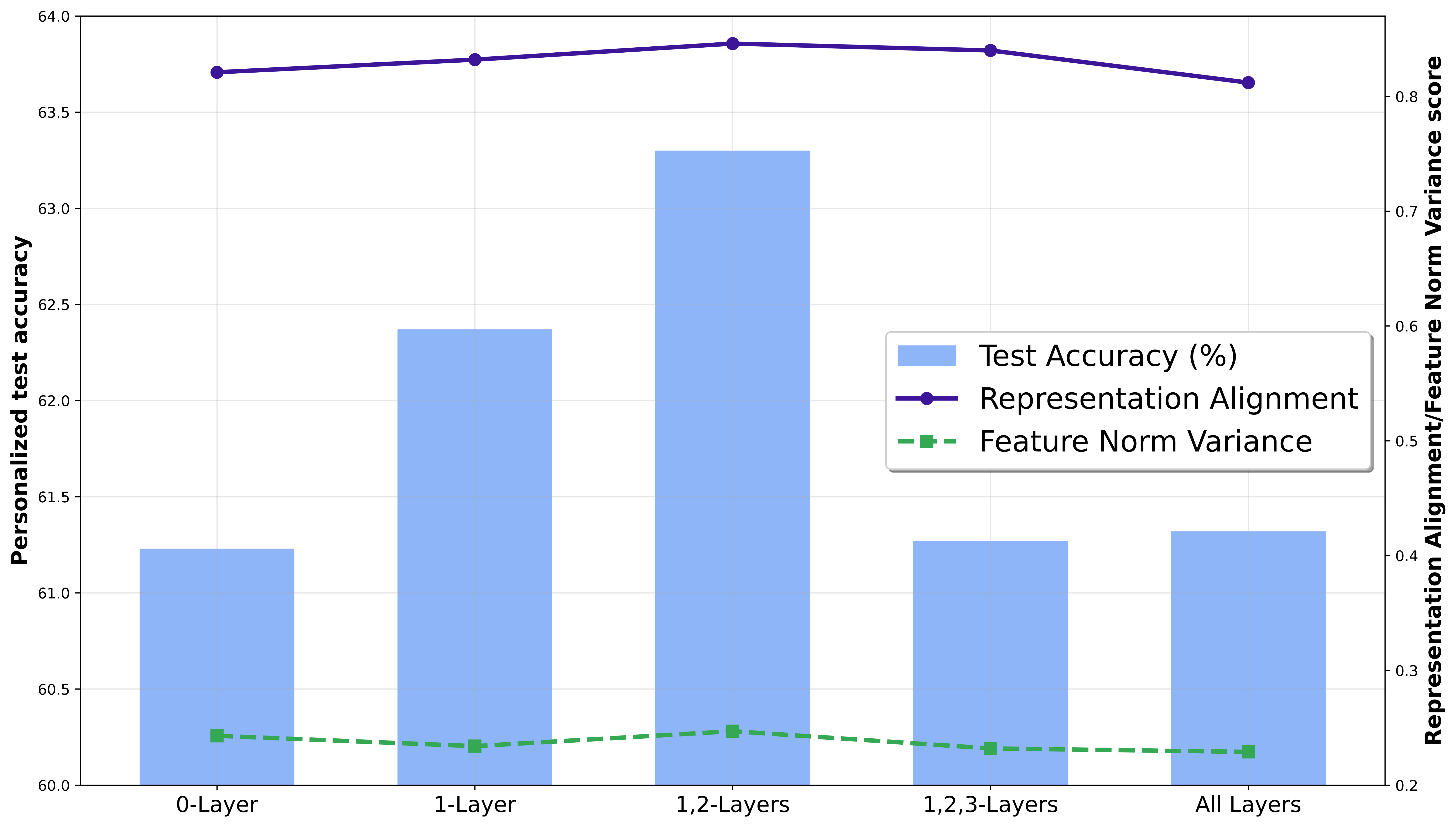}
    \caption{
        \textbf{Layer-wise ablation study of HDA.}
        We evaluate five configurations: a baseline using MSE loss (without cosine similarity) and settings with cosine similarity applied progressively from one to all layers. 
        The results show that increasing the depth of directional alignment leads to consistent improvements in test accuracy (blue bars), representation alignment (purple line), and feature norm variance (green dashed line).
    }
    \label{fig:HDA_ablation}
\end{figure}

As shown in Figure~\ref{fig:HDA_ablation}, personalized test accuracies steadily rise from the 0-layer to the 2-layer configuration, achieving their highest performance with the 2-layer setup. Both alignment and diversity scores show similar trends, indicating that the HDA configuration yields optimal alignment between personalized and global models.

\begin{figure}[t!]
    \centering
    \subcaptionbox{\label{fig:ablation4_per}Personalized test performance.}[0.49\linewidth]{
        \includegraphics[width=\linewidth,height=4.5cm,keepaspectratio]{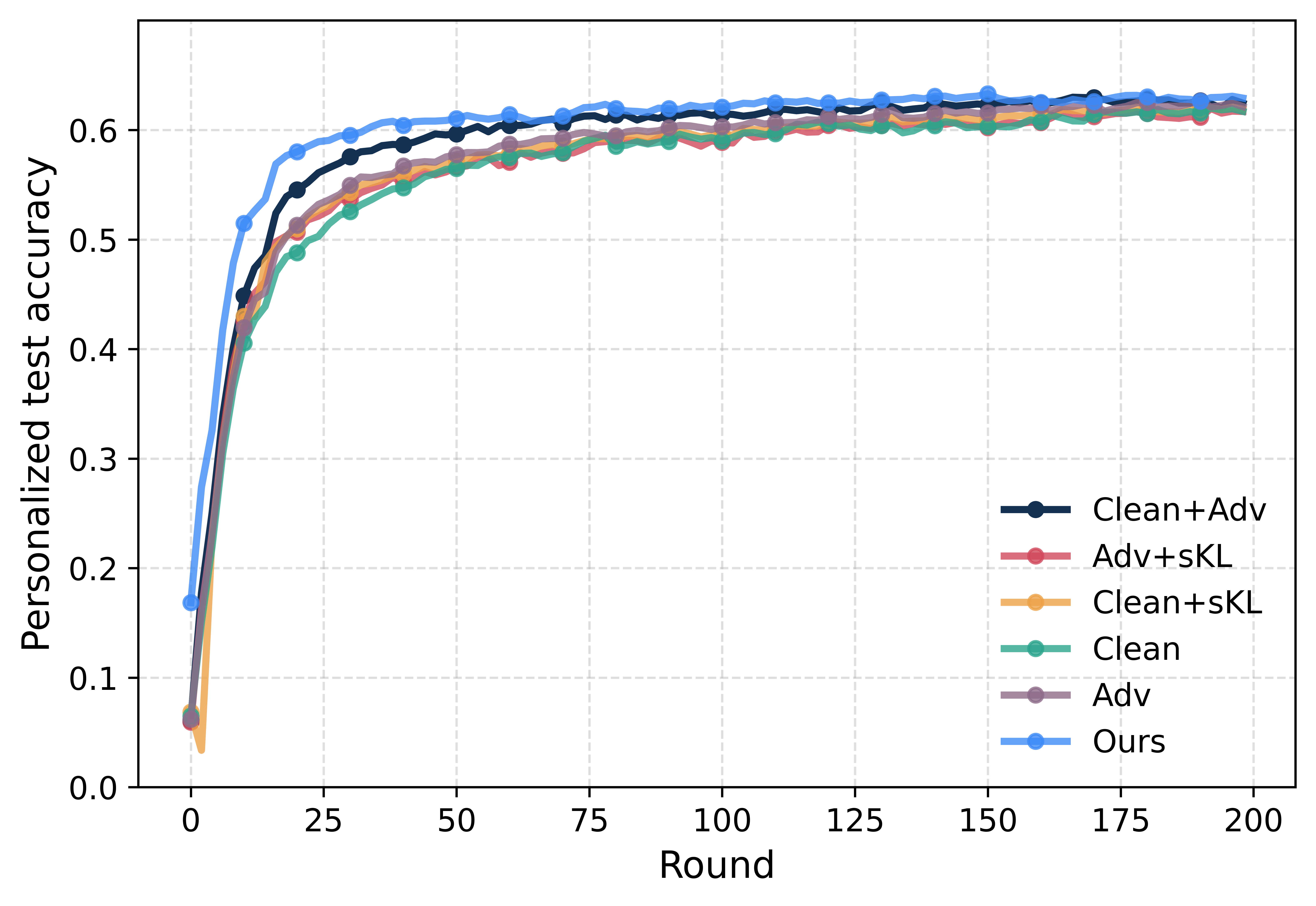}}
    \hfill
    \subcaptionbox{\label{fig:ablation4_server}Global test performance.}[0.49\linewidth]{
        \includegraphics[width=\linewidth,height=4.5cm,keepaspectratio]{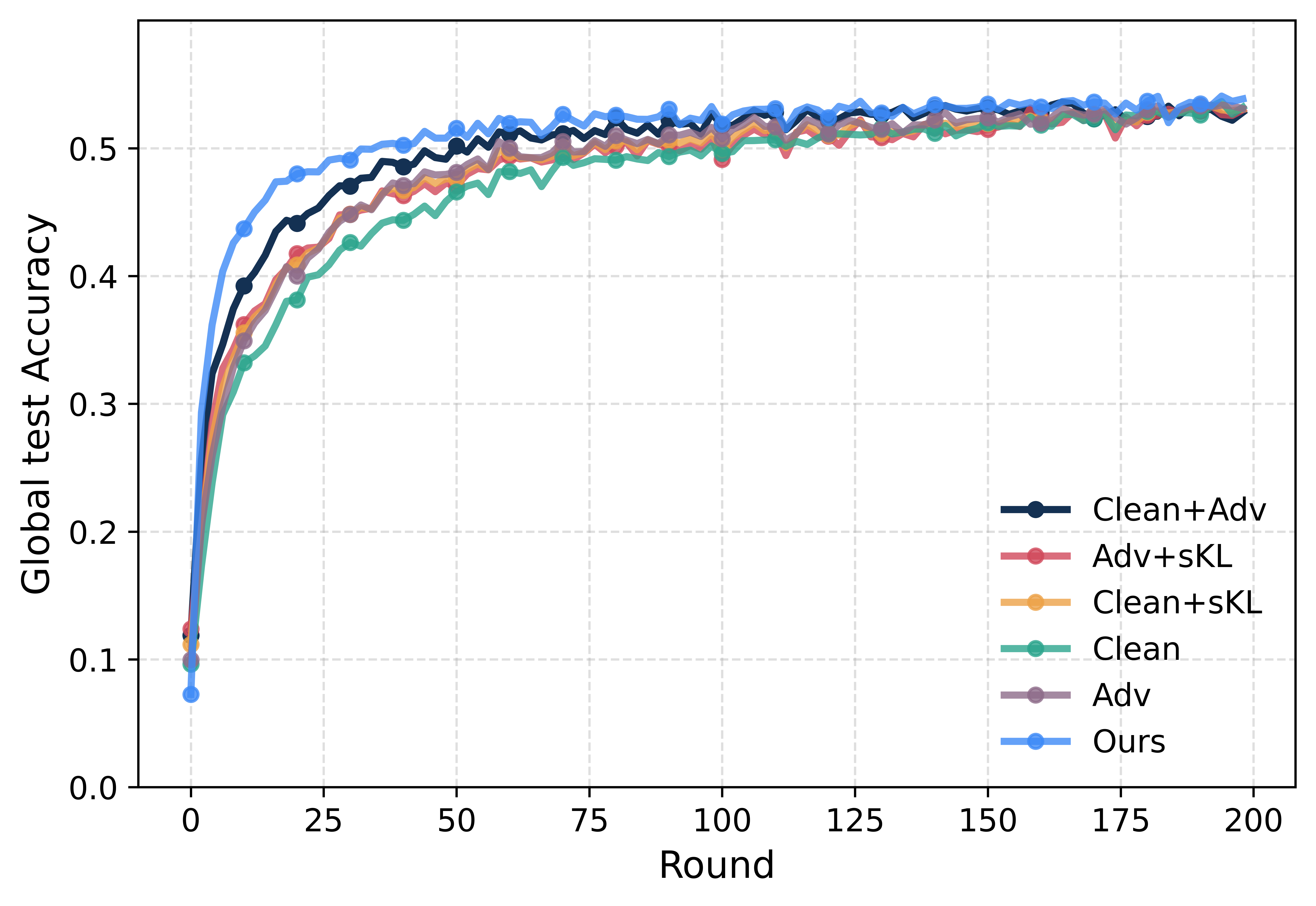}}
    \caption{Personalized and global test performance under component ablations of AKT on CIFAR100 with Dirichlet partitions ($\alpha=0.1$). We compare the AKT (Ours) against variants using only clean samples (Clean), using only adversarial perturbation (Adv), without adversarial perturbation but with symmetric KL (Clean+sKL), without symmetric KL but with adversarial perturbation (Clean+Adv), and without clean samples (Adv+sKL). Across both metrics, the Full AKT configuration achieves the best accuracy.}
\end{figure}

\noindent\textbf{Component Ablations in AKT.} We evaluate how each core component of AKT contributes to performance on CIFAR100 under Dirichlet client partitions with $\alpha=0.1$. Figure~\ref{fig:ablation4_per} and Figure~\ref{fig:ablation4_server} summarize the personalized and global test accuracies under these ablations.

Results show consistent patterns. First, our AKT outperforms all ablations on both metrics, indicating complementary gains from adversarial perturbation and symmetric KL. In particular, personalized test accuracy reaches 63.42\%, the highest among all variants, and global test accuracy reaches 54.08\%, also the best. Finally, the gains manifest in both evaluation regimes, indicating that the improvements are not confined to per-client adaptation but also propagate to the aggregated model. Taken together, these results substantiate that AKT is well-suited for PFL, delivering consistent benefits to both personalized and global performance.

\begin{figure}[t!]
    \centering

    \begin{subfigure}[b]{0.48\linewidth} 
        \centering
        \includegraphics[width=\linewidth]{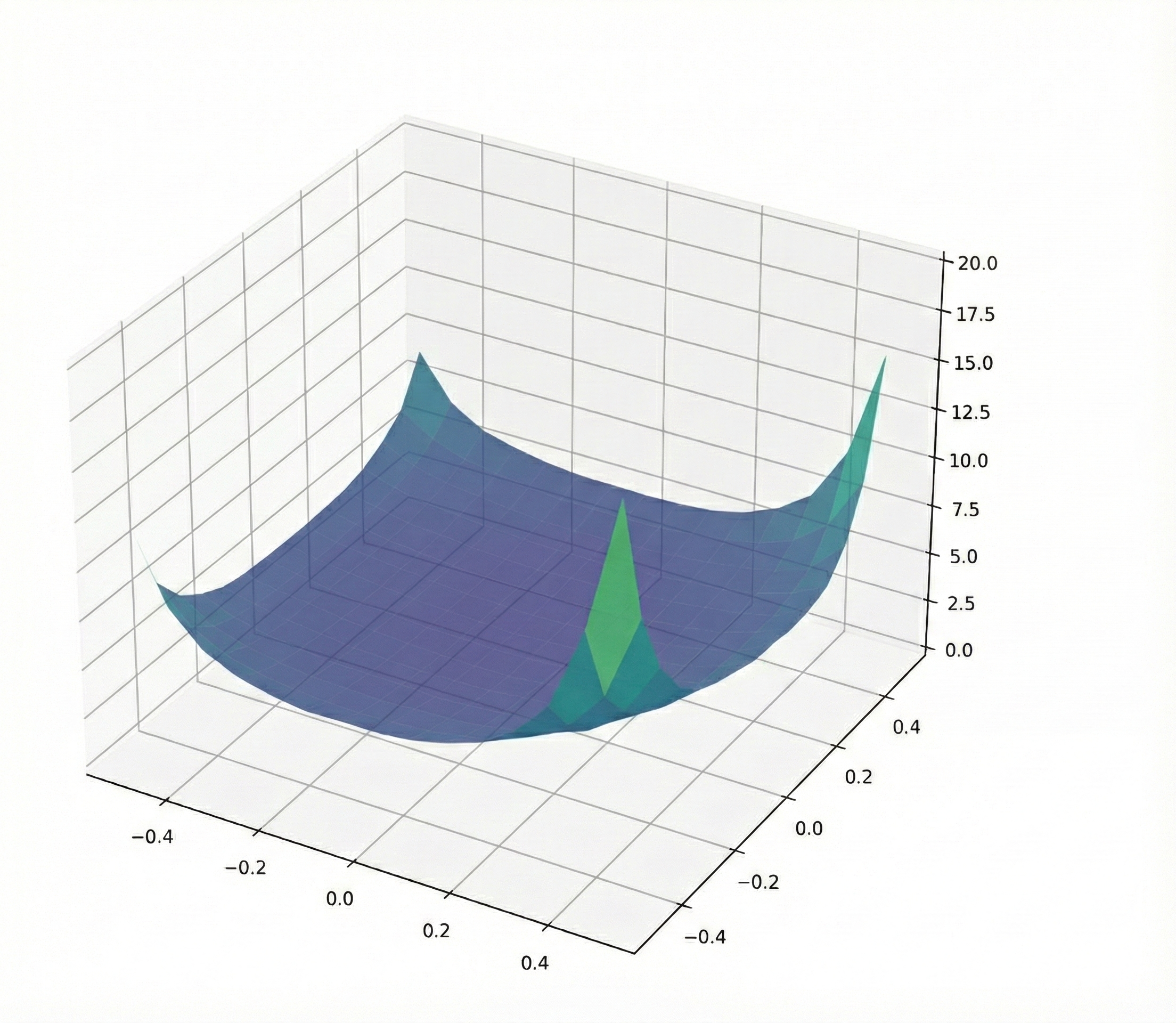}
        \caption{Clean}
        \label{fig:clean}
    \end{subfigure}
    \hfill 
    \begin{subfigure}[b]{0.48\linewidth}
        \centering
        \includegraphics[width=\linewidth]{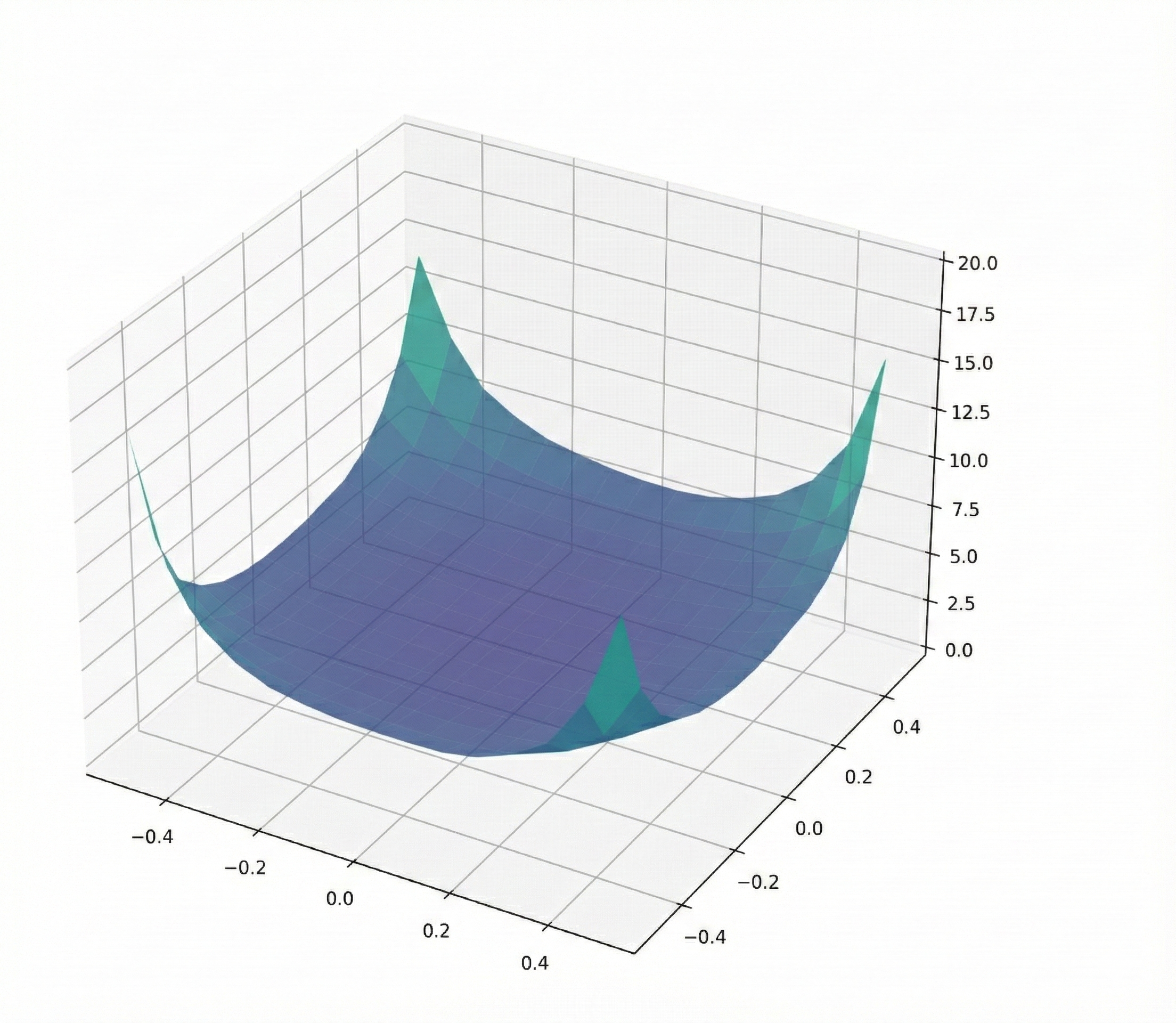}
        \caption{Adv}
        \label{fig:adv}
    \end{subfigure}


    \begin{subfigure}[b]{0.48\linewidth}
        \centering
        \includegraphics[width=\linewidth]{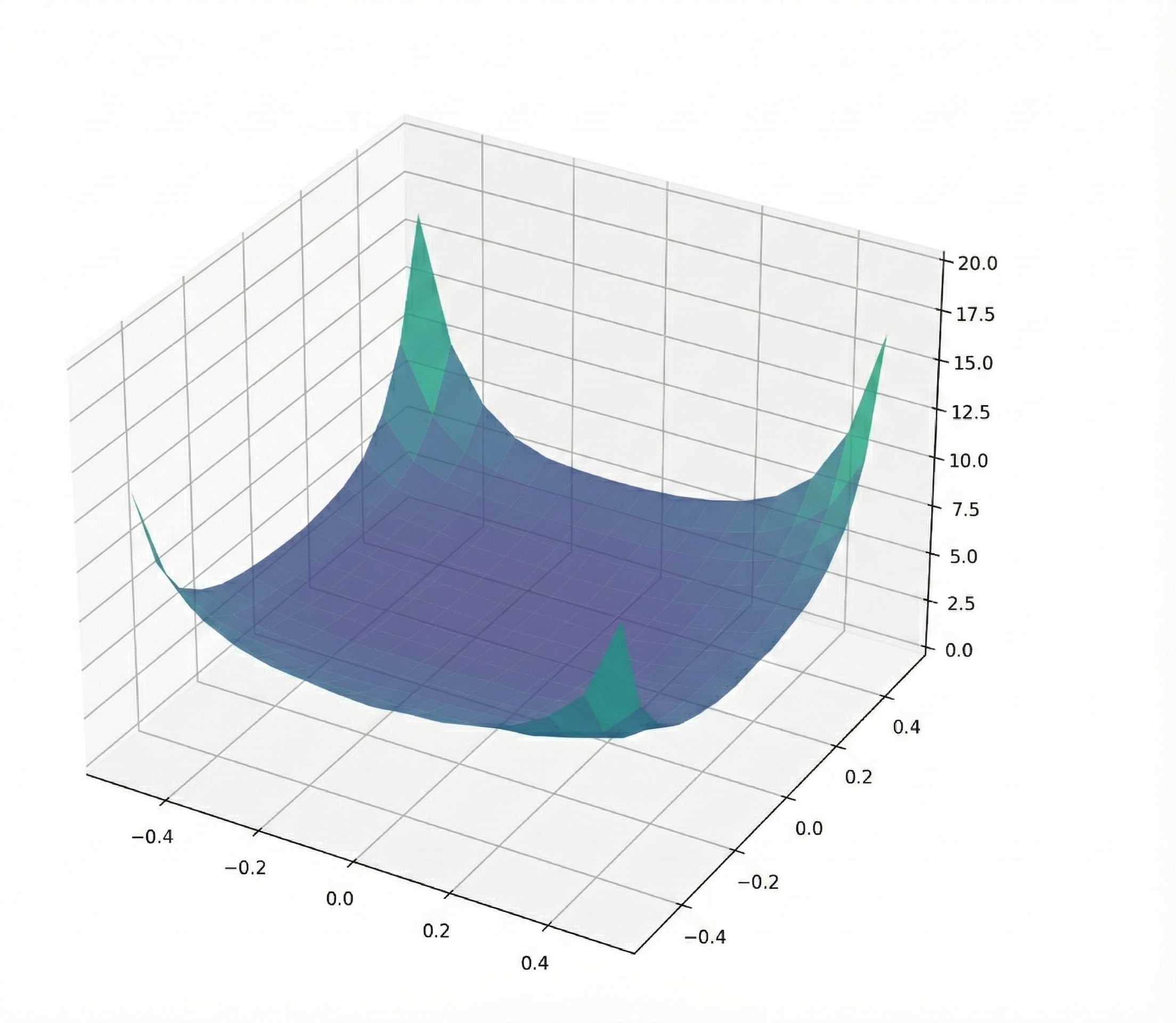}
        \caption{Clean + Adv}
        \label{fig:clean_adv}
    \end{subfigure}
    \hfill 
    \begin{subfigure}[b]{0.48\linewidth}
        \centering
        \includegraphics[width=\linewidth]{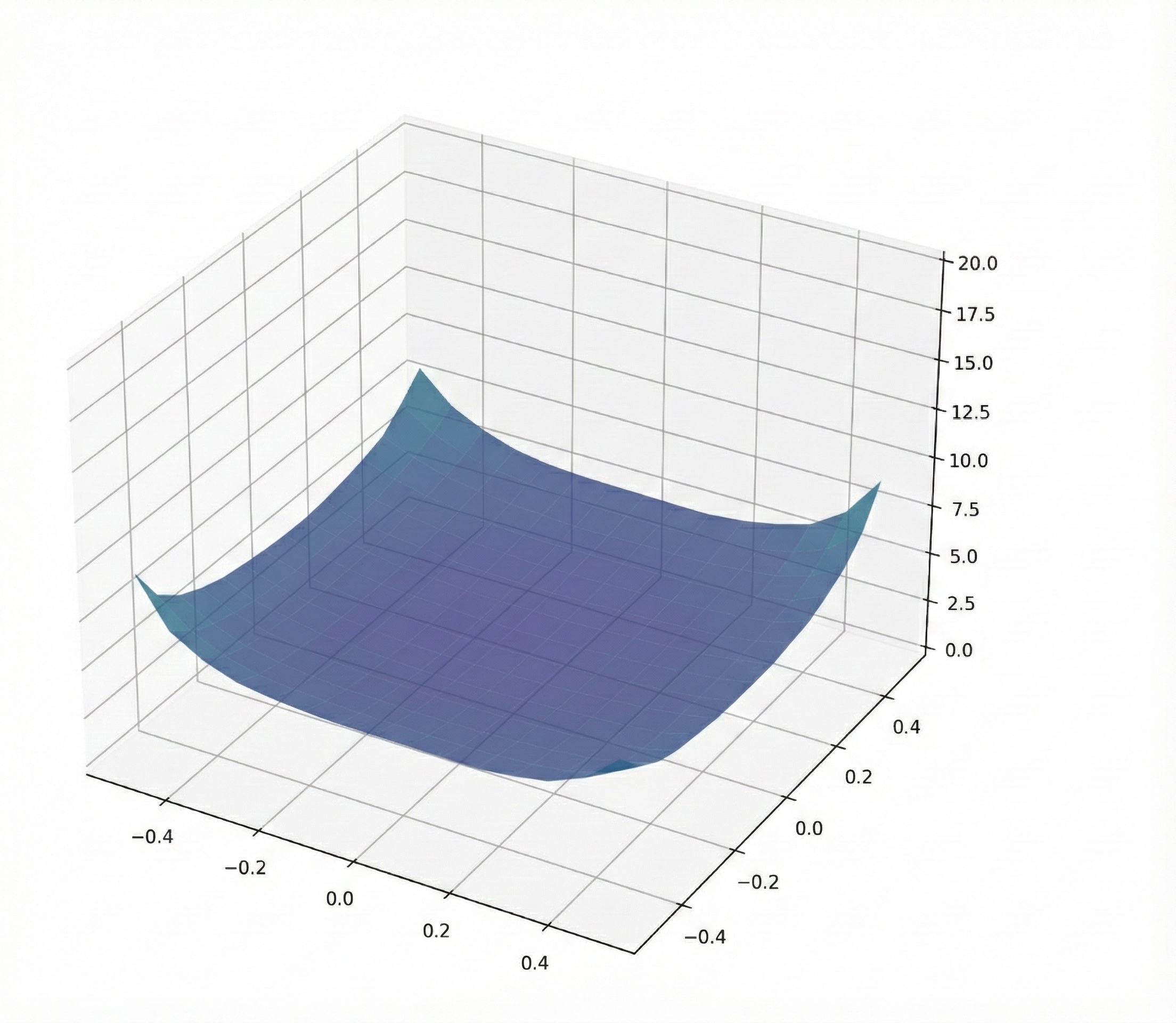}
        \caption{AKT (HEART-PFL)}
        \label{fig:ours}
    \end{subfigure}

    \caption{Visualization of the 3D loss landscapes for the four proxy sample scenarios. Each surface is generated by evaluating the average loss around the converged model parameters along two random normalized directions.}
    \label{fig:loss_landscape_ablation_3d}
\end{figure}

To verify that our proxy sample scenario facilitates the optimization stability underlying these performance gains, we examine the 3D loss surfaces of the four ablation variants (Fig.~\ref{fig:loss_landscape_ablation_3d}). This demonstrates that AKT effectively stabilizes aggregation and reduces bias across heterogeneous clients by reshaping the optimization landscape into a flatter region.

%% file: sec/5_conclusion.tex
\section{Conclusion}
We propose HEART-PFL to address the limitations of shallow feature alignment and fragile distillation in PFL. On the client side, HDA performs hierarchical alignment by applying cosine similarity to features of the early stage and MSE to the deep stage, preserving client-specific semantics. On the server side, AKT employs symmetric KL distillation on both clean and adversarial proxy samples to stabilize global updates. Experiments on CIFAR100, Flowers102, and Caltech101 show that HEART-PFL surpasses strong baselines with only 1.46M trainable parameters while maintaining robustness under out-of-domain proxy data. Ablation studies further demonstrate that HDA and AKT provide complementary benefits in alignment, robustness, and optimization stability. We conclude that HEART-PFL establishes a principled foundation for robust PFL under heterogeneity.

%% file: main.bib
@String(AAAI = {AAAI})

@article{jin2022personalized,
  title={Personalized edge intelligence via federated self-knowledge distillation},
  author={Jin, Hai and Bai, Dongshan and Yao, Dezhong and Dai, Yutong and Gu, Lin and Yu, Chen and Sun, Lichao},
  journal={IEEE Transactions on Parallel and Distributed Systems},
  volume={34},
  number={2},
  pages={567--580},
  year={2022},
  publisher={IEEE}
}

@techreport{krizhevsky2009learning,
  title       = {Learning Multiple Layers of Features from Tiny Images},
  author      = {Krizhevsky, Alex},
  institution = {University of Toronto},
  year        = {2009}
}

@inproceedings{li2020fedprox,
  title={Federated optimization in heterogeneous networks},
  author={Li, Tian and Sahu, Anit Kumar and Zaheer, Manzil and Sanjabi, Maziar and Talwalkar, Ameet and Smith, Virginia},
  booktitle={Proceedings of Machine learning and systems},
  volume={2},
  pages={429--450},
  year={2020}
}

@inproceedings{lin2020feddf,
  title={Ensemble distillation for robust model fusion in federated learning},
  author={Lin, Tao and Kong, Lingjing and Stich, Sebastian U and Jaggi, Martin},
  booktitle={Advances in Neural Information Processing Systems},
  volume={33},
  pages={2351--2363},
  year={2020}
}

@inproceedings{karimireddy2020scaffold,
  title={Scaffold: Stochastic controlled averaging for federated learning},
  author={Karimireddy, Sai Praneeth and Kale, Satyen and Mohri, Mehryar and Reddi, Sashank and Stich, Sebastian and Suresh, Ananda Theertha},
  booktitle={International conference on machine learning},
  pages={5132--5143},
  year={2020},
  organization={PMLR}
}

@inproceedings{t2020pfedme,
  title={Personalized federated learning with Moreau envelopes},
  author={T Dinh, Canh and Tran, Nguyen and Nguyen, Josh},
  booktitle={Advances in Neural Information Processing Systems},
  volume={33},
  pages={21394--21405},
  year={2020}
}

@inproceedings{arivazhagan2019fedper,
  title={Federated learning with personalization layers},
  author={Arivazhagan, Manoj Ghuhan and Aggarwal, Vinay and Singh, Ankit Kumar and Choudhury, Sunav},
  journal={arXiv preprint arXiv:1912.00818},
  year={2019}
}

@inproceedings{zhang2023gpfl,
  title={Gpfl: Simultaneously learning global and personalized feature information for personalized federated learning},
  author={Zhang, Jianqing and Hua, Yang and Wang, Hao and Song, Tao and Xue, Zhengui and Ma, Ruhui and Cao, Jian and Guan, Haibing},
  booktitle={Proceedings of the IEEE/CVF international conference on computer vision},
  pages={5041--5051},
  year={2023}
}

@inproceedings{li2021fedphp,
  title={Fedphp: Federated personalization with inherited private models},
  author={Li, Xin-Chun and Zhan, De-Chuan and Shao, Yunfeng and Li, Bingshuai and Song, Shaoming},
  booktitle={Joint European conference on machine learning and knowledge discovery in databases},
  pages={587--602},
  year={2021},
  organization={Springer}
}

@article{russakovsky2015imagenet,
  title={Imagenet large scale visual recognition challenge},
  author={Russakovsky, Olga and Deng, Jia and Su, Hao and Krause, Jonathan and Satheesh, Sanjeev and Ma, Sean and Huang, Zhiheng and Karpathy, Andrej and Khosla, Aditya and Bernstein, Michael and others},
  journal={International journal of computer vision},
  volume={115},
  number={3},
  pages={211--252},
  year={2015},
  publisher={Springer}
}

@article{fei2007learning,
  title={Learning generative visual models from few training examples: An incremental Bayesian approach tested on 101 object categories},
  author={Fei-Fei, Li and Fergus, Rob and Perona, Pietro},
  journal={Computer vision and Image understanding},
  volume={106},
  number={1},
  pages={59--70},
  year={2007},
  publisher={Elsevier}
}

@article{li2020federated,
  title={Federated optimization in heterogeneous networks},
  author={Li, Tian and Sahu, Anit Kumar and Zaheer, Manzil and Sanjabi, Maziar and Talwalkar, Ameet and Smith, Virginia},
  journal={Proceedings of Machine learning and systems},
  volume={2},
  pages={429--450},
  year={2020}
}

@inproceedings{tan2022fedproto,
  title={FedProto: Federated Prototype Learning across Heterogeneous Clients},
  author={Tan, Yue and Long, Guodong and Liu, Lu and Zhou, Tianyi and Lu, Qinghua and Jiang, Jing and Zhang, Chengqi},
  booktitle={Proceedings of the AAAI Conference on Artificial Intelligence},
  volume={36},
  number={8},
  pages={8432--8440},
  year={2022}
}

@article{lee2023self,
  title={Self-knowledge distillation via dropout},
  author={Lee, Hyoje and Park, Yeachan and Seo, Hyun and Kang, Myungjoo},
  journal={Computer Vision and Image Understanding},
  volume={233},
  pages={103720},
  year={2023},
  publisher={Elsevier}
}

@article{qi2025balance,
  title={Balance Divergence for Knowledge Distillation},
  author={Qi, Yafei and Wang, Chen and Zhang, Zhaoning and Liu, Yaping and Zhang, Yongmin},
  journal={arXiv preprint arXiv:2501.07804},
  year={2025}
}

@article{wu2024rethinking,
  title={Rethinking kullback-leibler divergence in knowledge distillation for large language models},
  author={Wu, Taiqiang and Tao, Chaofan and Wang, Jiahao and Yang, Runming and Zhao, Zhe and Wong, Ngai},
  journal={arXiv preprint arXiv:2404.02657},
  year={2024}
}

@article{zhang2021parameterized,
  title={Parameterized knowledge transfer for personalized federated learning},
  author={Zhang, Jie and Guo, Song and Ma, Xiaosong and Wang, Haozhao and Xu, Wenchao and Wu, Feijie},
  journal={Advances in Neural Information Processing Systems},
  volume={34},
  pages={10092--10104},
  year={2021}
}

@article{hanzely2020federated,
  title={Federated learning of a mixture of global and local models},
  author={Hanzely, Filip and Richt{\'a}rik, Peter},
  journal={arXiv preprint arXiv:2002.05516},
  year={2020}
}

@inproceedings{nilsback2008automated,
  title     = {Automated Flower Classification over a Large Number of Classes},
  author    = {Nilsback, Maria-Elena and Zisserman, Andrew},
  booktitle = {Proceedings of the Indian Conference on Computer Vision, Graphics and Image Processing (ICVGIP)},
  year      = {2008}
}

@inproceedings{tamirisa2024fedselect,
  title={Fedselect: Personalized federated learning with customized selection of parameters for fine-tuning},
  author={Tamirisa, Rishub and Xie, Chulin and Bao, Wenxuan and Zhou, Andy and Arel, Ron and Shamsian, Aviv},
  booktitle={Proceedings of the IEEE/CVF Conference on Computer Vision and Pattern Recognition},
  pages={23985--23994},
  year={2024}
}

@inproceedings{zhang2023fedcp,
  title={Fedcp: Separating feature information for personalized federated learning via conditional policy},
  author={Zhang, Jianqing and Hua, Yang and Wang, Hao and Song, Tao and Xue, Zhengui and Ma, Ruhui and Guan, Haibing},
  booktitle={Proceedings of the 29th ACM SIGKDD conference on knowledge discovery and data mining},
  pages={3249--3261},
  year={2023}
}

@inproceedings{zhang2023fedala,
  title={Fedala: Adaptive local aggregation for personalized federated learning},
  author={Zhang, Jianqing and Hua, Yang and Wang, Hao and Song, Tao and Xue, Zhengui and Ma, Ruhui and Guan, Haibing},
  booktitle={Proceedings of the AAAI conference on artificial intelligence},
  volume={37},
  number={9},
  pages={11237--11244},
  year={2023}
}

@article{oh2021fedbabu,
  title={Fedbabu: Towards enhanced representation for federated image classification},
  author={Oh, Jaehoon and Kim, Sangmook and Yun, Se-Young},
  journal={arXiv preprint arXiv:2106.06042},
  year={2021}
}

@inproceedings{collins2021exploiting,
  title={Exploiting shared representations for personalized federated learning},
  author={Collins, Liam and Hassani, Hamed and Mokhtari, Aryan and Shakkottai, Sanjay},
  booktitle={International conference on machine learning},
  pages={2089--2099},
  year={2021},
  organization={PMLR}
}

@inproceedings{pillutla2022federated,
  title={Federated learning with partial model personalization},
  author={Pillutla, Krishna and Malik, Kshitiz and Mohamed, Abdel-Rahman and Rabbat, Mike and Sanjabi, Maziar and Xiao, Lin},
  booktitle={International Conference on Machine Learning},
  pages={17716--17758},
  year={2022},
  organization={PMLR}
}

@article{liang2020think,
  title={Think locally, act globally: Federated learning with local and global representations},
  author={Liang, Paul Pu and Liu, Terrance and Ziyin, Liu and Salakhutdinov, Ruslan and Morency, Louis-Philippe},
  journal={arXiv preprint arXiv:2001.01523},
  year={2020}
}

@inproceedings{li2021ditto,
  title={Ditto: Fair and robust federated learning through personalization},
  author={Li, Tian and Hu, Shengyuan and Beirami, Ahmad and Smith, Virginia},
  booktitle={International conference on machine learning},
  pages={6357--6368},
  year={2021},
  organization={PMLR}
}

@article{hanzely2020lower,
  title={Lower bounds and optimal algorithms for personalized federated learning},
  author={Hanzely, Filip and Hanzely, Slavom{\'\i}r and Horv{\'a}th, Samuel and Richt{\'a}rik, Peter},
  journal={Advances in Neural Information Processing Systems},
  volume={33},
  pages={2304--2315},
  year={2020}
}

@article{mansour2020three,
  title={Three approaches for personalization with applications to federated learning},
  author={Mansour, Yishay and Mohri, Mehryar and Ro, Jae and Suresh, Ananda Theertha},
  journal={arXiv preprint arXiv:2002.10619},
  year={2020}
}

@article{t2020personalized,
  title={Personalized federated learning with moreau envelopes},
  author={T Dinh, Canh and Tran, Nguyen and Nguyen, Josh},
  journal={Advances in neural information processing systems},
  volume={33},
  pages={21394--21405},
  year={2020}
}

@inproceedings{gao2022feddc,
  title={Feddc: Federated learning with non-iid data via local drift decoupling and correction},
  author={Gao, Liang and Fu, Huazhu and Li, Li and Chen, Yingwen and Xu, Ming and Xu, Cheng-Zhong},
  booktitle={Proceedings of the IEEE/CVF conference on computer vision and pattern recognition},
  pages={10112--10121},
  year={2022}
}

@inproceedings{xiong2023feddm,
  title={Feddm: Iterative distribution matching for communication-efficient federated learning},
  author={Xiong, Yuanhao and Wang, Ruochen and Cheng, Minhao and Yu, Felix and Hsieh, Cho-Jui},
  booktitle={Proceedings of the IEEE/CVF Conference on Computer Vision and Pattern Recognition},
  pages={16323--16332},
  year={2023}
}

@inproceedings{duan2023federated,
  title={Federated learning with data-agnostic distribution fusion},
  author={Duan, Jian-hui and Li, Wenzhong and Zou, Derun and Li, Ruichen and Lu, Sanglu},
  booktitle={Proceedings of the IEEE/CVF Conference on Computer Vision and Pattern Recognition},
  pages={8074--8083},
  year={2023}
}

@article{li2019convergence,
  title={On the convergence of fedavg on non-iid data},
  author={Li, Xiang and Huang, Kaixuan and Yang, Wenhao and Wang, Shusen and Zhang, Zhihua},
  journal={arXiv preprint arXiv:1907.02189},
  year={2019}
}

@inproceedings{mcmahan2017communication,
  title={Communication-efficient learning of deep networks from decentralized data},
  author={McMahan, Brendan and Moore, Eider and Ramage, Daniel and Hampson, Seth and y Arcas, Blaise Aguera},
  booktitle={Artificial intelligence and statistics},
  pages={1273--1282},
  year={2017},
  organization={PMLR}
}

@article{madry2017towards,
  title={Towards deep learning models resistant to adversarial attacks},
  author={Madry, Aleksander and Makelov, Aleksandar and Schmidt, Ludwig and Tsipras, Dimitris and Vladu, Adrian},
  journal={arXiv preprint arXiv:1706.06083},
  year={2017}
}

@article{hinton2015distilling,
  title={Distilling the knowledge in a neural network},
  author={Hinton, Geoffrey and Vinyals, Oriol and Dean, Jeff},
  journal={arXiv preprint arXiv:1503.02531},
  year={2015}
}

@inproceedings{xie2024perada,
  title={Perada: Parameter-efficient federated learning personalization with generalization guarantees},
  author={Xie, Chulin and Huang, De-An and Chu, Wenda and Xu, Daguang and Xiao, Chaowei and Li, Bo and Anandkumar, Anima},
  booktitle={Proceedings of the IEEE/CVF conference on computer vision and pattern recognition},
  pages={23838--23848},
  year={2024}
}

@article{lin2020ensemble,
  title={Ensemble distillation for robust model fusion in federated learning},
  author={Lin, Tao and Kong, Lingjing and Stich, Sebastian U and Jaggi, Martin},
  journal={Advances in neural information processing systems},
  volume={33},
  pages={2351--2363},
  year={2020}
}

@inproceedings{chen2020simple,
  title={A simple framework for contrastive learning of visual representations},
  author={Chen, Ting and Kornblith, Simon and Norouzi, Mohammad and Hinton, Geoffrey},
  booktitle={International conference on machine learning},
  pages={1597--1608},
  year={2020},
  organization={PMLR}
}

@inproceedings{li2021model,
  title={Model-contrastive federated learning},
  author={Li, Qinbin and He, Bingsheng and Song, Dawn},
  booktitle={Proceedings of the IEEE/CVF international conference on computer vision},
  pages={10713--10722},
  year={2021}
}

@inproceedings{park2023understanding,
  title={Understanding the feature norm for out-of-distribution detection},
  author={Park, Jaewoo and Chun, Sae-Young and Kwak, Nojun},
  booktitle={Proceedings of the IEEE/CVF international conference on computer vision},
  pages={21544--21554},
  year={2023}
}

@article{liu2024fedgpd,
  title={Global prototype distillation for heterogeneous federated learning},
  author={Liu, Shaoming and Zhang, Yunfeng and others},
  journal={Scientific Reports},
  volume={14},
  pages={12284},
  year={2024},
  publisher={Nature}
}

@article{guo2025fedorgp,
  title={FedORGP: Guiding Heterogeneous Federated Learning with Orthogonality Regularization on Global Prototypes},
  author={Guo, Fucheng and Luan, Zeyu and Li, Qing and Zhao, Dan and Jiang, Yong},
  journal={arXiv preprint arXiv:2502.16119},
  year={2025}
}

@article{yao2024fedgkd,
  title={FedGKD: Global Knowledge Distillation in Federated Learning},
  author={Yao, Dawei and Zhou, Chang and Chew, Botong and Liang, Xinyu},
  journal={IEEE Transactions on Neural Networks and Learning Systems},
  volume={35},
  number={4},
  pages={4717--4729},
  year={2024},
  publisher={IEEE}
}

@article{li2019fedmd,
  title={FedMD: Heterogeneous Federated Learning via Model Distillation},
  author={Li, Dali and Wang, Junpu},
  journal={arXiv preprint arXiv:1910.03581},
  year={2019}
}
